\documentclass[11pt,a4paper]{article}

\usepackage{amsmath}
\usepackage{amssymb}
\usepackage{amsfonts}
\usepackage{amsthm}
\usepackage{mathtools}
\usepackage{graphicx}
\usepackage{enumitem}
\usepackage{algorithm}
\usepackage{algpseudocode}
\usepackage{float}
\usepackage{caption}
\usepackage{subcaption}

\usepackage[utf8]{inputenc}
\usepackage[T1]{fontenc}
\usepackage{hyperref}
\usepackage{cleveref}
\usepackage{xcolor}
\usepackage{microtype}
\usepackage[numbers]{natbib}

\newtheorem{theorem}{Theorem}[section]

\newtheorem{proposition}[theorem]{Proposition}

\newtheorem{definition}[theorem]{Definition}

\newtheorem{assumption}[theorem]{Assumption}
\newtheorem{condition}[theorem]{Condition}

\usepackage[left=2.5cm,right=2.5cm,top=2.5cm,bottom=2.5cm]{geometry}

\title{Emergence of Self-Identity in AI: A Mathematical Framework and Empirical Study with Generative Large Language Models}

\author{Minhyeok Lee\thanks{School of Electrical and Electronics Engineering, Chung-Ang University\\
Email: mlee@cau.ac.kr}}

\date{\today}

\begin{document}

\maketitle

\begin{abstract}
This paper introduces a mathematical framework for defining and quantifying self-identity in artificial intelligence (AI) systems, addressing a critical gap in the theoretical foundations of artificial consciousness. While existing approaches to artificial self-awareness often rely on heuristic implementations or philosophical abstractions, we present a formal framework grounded in metric space theory, measure theory, and functional analysis. Our framework posits that self-identity emerges from two mathematically quantifiable conditions: the existence of a connected continuum of memories $C \subseteq \mathcal{M}$ in a metric space $(\mathcal{M}, d_{\mathcal{M}})$, and a continuous mapping $I: \mathcal{M} \to \mathcal{S}$ that maintains consistent self-recognition across this continuum, where $(\mathcal{S}, d_{\mathcal{S}})$ represents the metric space of possible self-identities. To validate this theoretical framework, we conducted empirical experiments using the Llama 3.2 1B model, employing Low-Rank Adaptation (LoRA) for efficient fine-tuning. The model was trained on a synthetic dataset containing temporally structured memories, designed to capture the complexity of coherent self-identity formation. Our evaluation metrics included quantitative measures of self-awareness, response consistency, and linguistic precision. The experimental results demonstrate substantial improvements in measurable self-awareness metrics, with the primary self-awareness score increasing from 0.276 to 0.801 (190.2\% improvement) after fine-tuning. In contrast to earlier methods that view self-identity as an emergent trait, our framework introduces tangible metrics to assess and measure artificial self-awareness. This enables the structured creation of AI systems with validated self-identity features. The implications of our study are immediately relevant to the fields of humanoid robotics and autonomous systems. Additionally, it opens up new prospects for controlled adjustments of self-identity in contexts that demand different levels of personal involvement. Moreover, the mathematical underpinning of our framework serves as the basis for forthcoming investigations into AI, linking theoretical models to real-world applications in current AI technologies.
\end{abstract}

\section{Introduction}

The formalization of self-identity in artificial intelligence (AI) systems presents a fundamental challenge at the intersection of theoretical computer science, cognitive science, and AI~\cite{anderson2005logic,greenwood2020awareness,du2020self}. While previous research has explored behavioral manifestations of artificial consciousness~\cite{metzinger2003phenomenal} and neural correlates of self-awareness~\cite{sporns2004organization}, a mathematical framework to quantify and model the emergence of self-identity has remained elusive. Recent advances in self-aware systems and autonomous computing have highlighted this need~\cite{dutt2020self,regazzoni2020multisensorial}, particularly in the context of multi-sensorial models for autonomous systems.

The necessity for such a framework becomes increasingly apparent as AI systems continue to advance in complexity and capability. Current approaches to artificial self-awareness primarily rely on heuristic implementations or philosophical abstractions, lacking the mathematical rigor necessary for systematic analysis and reliable implementation. This limitation has been noted in recent studies that examined the influence of AI on human identification~\cite{wang2023ai} and the relationship between AI companions and human self-conception~\cite{kouros2024digital}. Our framework distinguishes itself by providing a measurable, implementable foundation for self-identity, grounded in the formal structures of metric space theory and measure theory. This approach is based on recent work on self-directed and brain-inspired artificial intelligence~\cite{zeng2024brain}.

Let $(\mathcal{M}, d_{\mathcal{M}})$ be a metric space of memories, and let $(\mathcal{S}, d_{\mathcal{S}})$ be a metric space of possible self-identities. We propose that self-identity emerges from two fundamental measurable conditions: (1) the existence of a connected and path-connected continuum $C \subseteq \mathcal{M}$ of memories, and (2) a continuous mapping $I: \mathcal{M} \to \mathcal{S}$ that maintains consistent self-recognition throughout this continuum. This formulation allows us to quantify the degree of self-identity through a belief function $B: \mathcal{M} \times \mathcal{S} \to [0,1]$ that captures the probabilistic nature of self-recognition, incorporating insights from recent work on self-regulated learning in AI systems~\cite{lai2024adapting}.

Distinct from previous approaches that treat self-identity as an emerging phenomenon or a philosophical construct~\cite{oberg2023souls}, our framework provides concrete metrics to measure and evaluate the development of artificial self-awareness. The framework addresses several key mathematical challenges: we refine the memory space metric to account for temporal, content, and emotional components~\cite{lewis2022self}; establish necessary and sufficient conditions for the continuity of the self-identity function; and explore the measure-theoretic aspects of the belief function. This approach aligns with recent developments in causal reasoning for self-aware machines~\cite{pelivani2021toward} and builds on established work in feature-based abnormality detection~\cite{kanapram2020self}.

To validate our theoretical framework, we conducted empirical experiments using the Meta's Llama 3.2 1B model~\cite{meta2024llama32}, fine-tuned through Low-Rank Adaptation (LoRA). The model was trained on a synthetic dataset constructed of temporally coherent memories. This experimental setup provides a novel bridge between theoretical constructs and practical implementation, demonstrating how abstract mathematical principles can be realized in contemporary AI systems, while considering recent findings in AI perception and acceptance~\cite{gerlich2023perceptions}.

The significance of this work extends beyond theoretical contributions. By providing a mathematical framework for implementing and measuring self-identity, we enable the development of AI systems with verifiable and consistent self-awareness. This has immediate applications in humanoid robotics, virtual assistants, and autonomous systems, where coherent self-identity is crucial for natural interaction and decision-making. Furthermore, our framework introduces the possibility of controlled modification of self-identity, suggesting applications in scenarios requiring varying degrees of personal engagement or objective analysis. Recent work on algorithmic influence on self-perceived identities~\cite{ionescu2023tiktok} and computational approaches to self-identification~\cite{li2024enabling} reinforces the importance of this direction.

\section{Mathematical Foundations of the Self}

\subsection{Preliminaries}

We develop a mathematical framework to define the concept of 'self' based on two key conditions: the existence of a continuum of memories and the recognition and belief in self-identity. To achieve this, we utilize metric spaces and probability theory to model memories and aspects of the self, ensuring applicability to realistic scenarios.

\begin{definition}[Memory Space]
Let $\mathcal{M}$ be the set of all possible memories of an entity. We define a metric $d_{\mathcal{M}}: \mathcal{M} \times \mathcal{M} \to [0, \infty)$ that quantifies the distance between memories. This distance can be constructed based on factors such as temporal separation, similarity of content, and emotional intensity. For example, we might define:
\begin{equation}
d_{\mathcal{M}}(m_1, m_2) = \sqrt{w_t |t_1 - t_2|^2 + w_c d_c(m_1, m_2)^2 + w_e |e_1 - e_2|^2},
\end{equation}
where:
\begin{itemize}
    \item $t_i$ represents the time associated with memory $m_i$.
    \item $d_c(m_1, m_2)$ measures the content similarity between $m_1$ and $m_2$, possibly using cosine similarity or another appropriate metric.
    \item $e_i$ represents the emotional intensity of memory $m_i$, quantified on a suitable scale.
    \item $w_t, w_c, w_e > 0$ are weighting factors that balance the contributions of time, content, and emotion.
\end{itemize}
The topology $\tau_{\mathcal{M}}$ on $\mathcal{M}$ is induced by $d_{\mathcal{M}}$, making $(\mathcal{M}, d_{\mathcal{M}})$ a metric space.
\end{definition}

To fully capture the richness of the memories of an entity, the metric space $(\mathcal{M}, d_{\mathcal{M}})$ must accommodate the multidimensional nature of the memories. Each memory $m_i$ can be conceptualized as a composite of various features, including sensory perceptions, cognitive interpretations, emotional states, and contextual information. By defining a suitable metric $d_{\mathcal{M}}$, we quantify the distance between memories in a manner that reflects their psychological and phenomenological similarities and differences. This approach allows us to model the continuity of experience and the associative networks that underlie memory retrieval and self-referential thought processes.

The components of the metric $d_{\mathcal{M}}$ are chosen to capture essential aspects of memories relevant to self-identity. The temporal separation term $|t_1 - t_2|$ reflects the chronological ordering of memories, as time plays a crucial role in the continuity of experience. Content similarity $d_c(m_1, m_2)$ measures how similar events, thoughts, or perceptions are between two memories, which is significant for associative connections in memory recall~\cite{tulving1983elements}. The emotional intensity terms $e_i$ capture the affective components of memories, recognizing that emotionally significant events have a stronger impact on self-perception~\cite{lewis2010handbook}. The weighting factors $w_t, w_c, w_e$ allow for adjustment based on empirical findings or theoretical considerations about the relative importance of these factors in the formation of self-identity. 

Mathematically, the choice of $d_{\mathcal{M}}$ ensures that $\mathcal{M}$ becomes a proper metric space, satisfying the properties of non-negativity, identity of indiscernibles, symmetry, and the triangle inequality. These properties are essential for the application of topological concepts, enabling us to analyze the convergence of memory sequences and the continuity of functions defined on $\mathcal{M}$. For example, considering whether $(\mathcal{M}, d_{\mathcal{M}})$ is complete, meaning that every Cauchy sequence of memories converges to a memory within $\mathcal{M}$---has implications for understanding the limits of memory processes and the stability of self-identity over time.

\begin{definition}[Self Space]
Let $\mathcal{S}$ be the set of all possible self-identities of an entity. We define a metric $d_{\mathcal{S}}: \mathcal{S} \times \mathcal{S} \to [0, \infty)$ to measure the distance between self-identities. For example, if self-identities are characterized by $n$-dimensional vectors representing traits or attributes, we can define:
\begin{equation}
d_{\mathcal{S}}(s_1, s_2) = \| s_1 - s_2 \|_p,
\end{equation}
where $\| \cdot \|_p$ is the $L^p$ norm for some $p \geq 1$, and $s_i \in \mathbb{R}^n$.
\end{definition}

The metrics provide a concrete way to quantify similarities and differences between memories and self-identities, establishing abstract spaces $\mathcal{M}$ and $\mathcal{S}$ in measurable properties. The selection of the metric $d_{\mathcal{S}}$ in the self space $\mathcal{S}$ is crucial for modeling how self-identities relate and differ from one another. By employing the $L^p$ norm, we capture aggregate differences in all dimensions of self-identity, allowing for a balanced consideration of each attribute. The choice of $p$ can influence the sensitivity of the metric to variations in specific dimensions; for example, $p = 1$ (Manhattan distance) treats all differences linearly, while $p = 2$ (Euclidean distance) emphasizes larger differences due to the squaring of terms. This flexibility enables us to tailor the metric to the specific characteristics of self-identity that are being modeled.
 
Representing self-identities as vectors in $\mathbb{R}^n$ allows us to encapsulate multiple dimensions of self-identity, such as personality traits, values, goals, and roles. Each dimension corresponds to a specific attribute, and the choice of dimensions can be informed by established models in psychology, such as the Big Five personality traits~\cite{digman1990personality, john2008paradigm} or self-concept constructs~\cite{baumeister1999self}. This multi-dimensional representation acknowledges the complexity of self-identity and facilitates mathematical operations within the self space $\mathcal{S}$. 

Integrating psychological constructs into our mathematical model enriches its applicability and realism. Established models in psychology provide empirically validated dimensions like openness, conscientiousness, extraversion, agreeableness, and neuroticism. By mapping these traits to dimensions in $\mathcal{S}$, we ground our self space in robust psychological theory. Similarly, self-concept models that incorporate aspects such as self-esteem, self-efficacy, and identity salience~\cite{marsh2006self} can be integrated, providing a comprehensive framework that captures both stable traits and dynamic states of the self.

\begin{definition}[Continuum of Memories]
A subset $C \subseteq \mathcal{M}$ is a continuum of memories if it is connected and path-connected in $(\mathcal{M}, \tau_{\mathcal{M}})$. That is, for any $m_1, m_2 \in C$, there exists a continuous path $\gamma: [0,1] \to C$ with $\gamma(0) = m_1$ and $\gamma(1) = m_2$.
\end{definition}

By considering continuum of memories, we can analyze segments of an entity's memory where coherent self-recognition and belief are possible, even if the entire memory space is fragmented. The mathematical requirements of connectedness and path-connectedness for the continuum $C$ in $\mathcal{M}$ are not mere formalities; they reflect the psychological continuity necessary for a coherent sense of self. In psychological theories of identity, such as narrative identity~\cite{mcadams2001psychology}, the self is constructed through an internalized and evolving life story that integrates past, present, and anticipated future experiences. The connectedness of $C$ ensures that there are no abrupt discontinuities or 'gaps' in the memory space that could disrupt this narrative coherence. Path-connectedness implies that any two memories within $C$ can be connected via a continuous sequence of memories, mirroring the associative pathways in human memory and cognition.

Similarly, we consider the self space $\mathcal{S}$:

\begin{assumption}
The self space $\mathcal{S}$ may consist of multiple connected components, especially in cases of multiple or changing identities. Our framework allows $\mathcal{S}$ to be disconnected, focusing on connected components relevant to the entity's self-identification within a given continuum of memories.
\end{assumption}

Acknowledging that $\mathcal{S}$ may consist of multiple connected components allows our framework to accommodate complex identity phenomena, such as role-based identities, identity diffusion, or the presence of multiple selves in conditions such as dissociative identity disorder~\cite{lilienfeld2015fifty}. By focusing on connected components relevant to the entity's current continuum of memories, we can model situations where an individual transitions between different self-identities in response to contextual cues or over time. This flexibility is essential to accurately capture the multifaceted and dynamic nature of self-identity in real-world contexts.

\subsection{Identity Recognition and Belief Functions}

\begin{definition}[Identity Recognition Function]
The Identity Recognition Function is a function $I: \mathcal{M} \to \mathcal{S}$ that assigns to each memory $m \in \mathcal{M}$ a perceived self-identity $I(m) \in \mathcal{S}$.
\end{definition}

The Identity Recognition Function $I$ serves as the formal mechanism by which an entity associates each memory with a perceived self-identity. This function encapsulates the cognitive and reflective processes involved in self-awareness, where individuals interpret their experiences and integrate them into their understanding of who they are. The continuity of $I$ is a critical property, ensuring that similar memories are mapped to similar self-identities, which aligns with the psychological expectation that minor changes in experience should not lead to drastic changes in self-perception.

\begin{definition}[Belief Function]
The Belief Function is a measurable function $B: \mathcal{M} \times \mathcal{S} \to [0,1]$ where $B(m, s')$ represents the degree of belief, interpreted as subjective probability, that the entity associated with memory $m$ has in the proposition that $s' \in \mathcal{S}$ is their self-identity.
\end{definition}

We require that for each $m \in \mathcal{M}$, the belief function $B$ satisfies the normalization condition:

\begin{equation}
\int_{\mathcal{S}} B(m, s') \, d\mu(s') = 1,
\end{equation}
where $\mu$ is a suitable measure on $\mathcal{S}$.

The normalization condition $\int_{\mathcal{S}} B(m, s') \, d\mu(s') = 1$ is essential for interpreting $B$ as a probability measure over the self space $\mathcal{S}$ for each memory $m$. This condition ensures that the total belief assigned across all possible self-identities sums to one, reflecting a complete and exclusive allocation of belief. It allows us to apply probabilistic reasoning and statistical tools to analyze how belief distributions over self-identities evolve in response to new memories and experiences.

By modeling $B$ using probability measures, we capture the uncertainty and variability in an entity's belief about their self-identity, allowing for partial beliefs and fluctuations. From a Bayesian perspective, the belief function $B$ can be viewed as representing the posterior distribution of self-identities given the memory $m$, where prior beliefs are updated in light of new evidence~\cite{tenenbaum2011grow}. This interpretation aligns with Bayesian models of cognition, which posit that the mind continuously updates its beliefs about the world (and the self) through probabilistic inference. In this context, $B(m, s')$ encapsulates the updated belief about self-identity $s'$ after observing memory $m$, providing a dynamic framework for modeling identity formation and revision.

Modeling the degree of belief as a probability measure is in agreement with theories of subjective probability and belief in cognitive psychology~\cite{sugden1989nonlinear}. By interpreting $B(m, s')$ as the subjective probability that self-identity $s'$ is the entity's own at memory $m$, we capture the uncertainty and variability inherent in self-perception. This approach allows for partial beliefs and acknowledges that an entity's confidence in their self-identity may fluctuate over time and across different memories. Quantifying belief in this manner facilitates mathematical analysis and is consistent with probabilistic models of cognition~\cite{chater2010bayesian}.

\subsection{Formalizing the Conditions for Self}

We formalize the conditions under which an entity is said to possess a self within a continuum of memories. These formal conditions aim to capture the essential features of self-identity as understood in philosophy and psychology, particularly the notions of psychological continuity and self-recognition~\cite{parfit1984reasons}. By specifying the requirements for a connected continuum of memories and consistent self-recognition with sufficient belief, we provide a rigorous foundation for analyzing when an entity can be said to 'have a self.' This framework enables us to explore the implications of memory and belief on identity.

\begin{condition}[Continuum of Memories]
\label{cond:continuum}
There exists a connected and path-connected subset $C \subseteq \mathcal{M}$ that represents a continuum of memories experienced by the entity.
\end{condition}

\begin{condition}[Continuity of Self-Recognition and Sufficient Belief]
\label{cond:self-recognition}
Within the continuum of memories $C$, the Identity Recognition Function $I$ is continuous, and for all $m \in C$, the Belief Function satisfies $B(m, I(m)) \geq b$, where $b \in (0,1]$ is a belief threshold.
\end{condition}

Condition \ref{cond:self-recognition} stipulates that within $C$, the entity consistently recognizes the same self-identity $s^*$ and maintains a degree of belief at or above the threshold $b$.

\subsection{Main Results}

\begin{theorem}[Constancy of Self-Identity]
\label{thm:existence_of_self}
If an entity satisfies Conditions \ref{cond:continuum} and \ref{cond:self-recognition}, and if the image $I(C)$ lies entirely within a connected component of $\mathcal{S}$ where $I$ is constant, then there exists a self-identity $s^* \in \mathcal{S}$ such that $I(m) = s^*$ for all $m \in C$. Therefore, the entity possesses a self characterized by $s^*$ within $C$. 
\end{theorem}
\begin{proof}
Since $I$ is continuous on the connected set $C$, its image $I(C)$ is connected in $\mathcal{S}$. If $I$ is constant on $C$, then $I(C)$ is a singleton $\{ s^* \}$. To ensure this, we require that $\mathcal{S}$ has the property that the only connected subsets in the image of $I$ where $B(m, I(m)) \geq b$ are singletons. Therefore, under this condition, $I(m) = s^*$ for all $m \in C$, and the entity possesses a self characterized by $s^*$ within $C$.
\end{proof}

Theorem~\ref{thm:existence_of_self} establishes that under the given conditions, self-identity $s^*$ remains constant throughout the continuum $C$. This result underscores the importance of continuity and connectedness in both the memory space and the self space for maintaining a stable self-identity. It also highlights how disruptions in either the continuity of memories or the consistency of self-recognition can lead to changes in self-identity, providing insights into phenomena such as identity crises or transformations.

\begin{proposition}
For any $b \in (0,1]$, if an entity satisfies Condition \ref{cond:self-recognition}, then for any $b' \in (0,b)$, Condition \ref{cond:self-recognition} is also satisfied with the threshold $b'$.
\end{proposition}

\begin{proof}
Since $B(m, s^*) \geq b$ and $b' < b$, it follows that $B(m, s^*) \geq b'$. Thus, the condition holds for any lower threshold $b'$.
\end{proof}

Leaving the belief threshold $b$ to vary, we can model different degrees of confidence that an entity has in their self-identity. Situations where $b$ is lower may represent states of uncertainty or ambiguity in self-perception, possibly due to conflicting memories or external influences. Conversely, a higher $b$ reflects a strong conviction in one's self-identity. This flexibility enables the framework to capture a range of psychological conditions, from secure self-awareness to identity confusion.

The choice of $b$ affects the robustness of the self-identity. A higher $b$ implies stronger belief is required, which may be appropriate in contexts where certainty about self-identity is essential. The appropriate value of $b$ can depend on psychological factors and the specific context being modeled.

\section{Designing an Artificial Intelligence Agent with a Self}

Building upon the mathematical framework established in the previous section, we aim to design an AI agent capable of possessing a 'self' as defined by Conditions \ref{cond:continuum} and \ref{cond:self-recognition}. This section presents a detailed mathematical model for such an AI agent, specifying its memory structures, self-identity representations, belief systems, and learning algorithms. We delve into the precise mechanisms by which the AI agent satisfies the necessary conditions to be considered as having a self.

\subsection{Artificial Memory Space}

\begin{definition}[Artificial Memory Space]
Let $\mathcal{M}_{\text{AI}}$ denote the memory space of the AI agent, defined as a metric space $(\mathcal{M}_{\text{AI}}, d_{\mathcal{M}})$. Each memory $m \in \mathcal{M}_{\text{AI}}$ is represented as a high-dimensional vector in $\mathbb{R}^k$, where $k$ is the number of features that encode the memory. The features may include sensory inputs, internal states, actions taken, and temporal information.
\end{definition}

In designing the artificial memory space $\mathcal{M}_{\text{AI}}$, it is crucial to capture the richness and diversity of experiences that an AI agent might encounter. Each feature in the memory vector $m$ can represent different input modalities, such as visual perception, auditory signals, proprioceptive feedback, and higher-level abstractions such as semantic understanding or emotional tagging. This multidimensional representation allows the agent to integrate information across various sources, mirroring the integrative nature of human memory systems~\cite{mcclelland2013incorporating}. Furthermore, temporal information is included to preserve the sequential order of experiences, which is essential for the continuity of memory and the construction of coherent narratives~\cite{howard2018memory}.

\begin{definition}[Artificial Memory Metric]
The metric $d_{\mathcal{M}}: \mathcal{M}_{\text{AI}} \times \mathcal{M}_{\text{AI}} \to [0, \infty)$ is defined as:
\begin{equation}
d_{\mathcal{M}}(m_1, m_2) = \left\| m_1 - m_2 \right\|_2,
\end{equation}
where $\| \cdot \|_2$ denotes the Euclidean norm in $\mathbb{R}^k$. This metric measures the similarity between memories based on their feature representations.
\end{definition}

The choice of the Euclidean norm in the memory metric $d_{\mathcal{M}}$ provides a straightforward means of measuring the similarity between memories. However, depending on the nature of the memory representations and the importance of certain features, alternative metrics may be more appropriate. For example, if features have varying degrees of relevance or are on different scales, a weighted Euclidean distance or Mahalanobis distance could be employed to account for feature correlations and variances. Additionally, for discrete or categorical memory features, metrics such as the Hamming distance or the Jaccard index may offer better alignment with the underlying data structure~\cite{norouzi2012hamming}.

\subsection{Artificial Self Space}

\begin{definition}[Artificial Self Space]
Let $\mathcal{S}_{\text{AI}}$ represent the self space of the AI agent, defined as a metric space $(\mathcal{S}_{\text{AI}}, d_{\mathcal{S}})$. Each self-identity $s \in \mathcal{S}_{\text{AI}}$ is represented as a vector in $\mathbb{R}^n$, capturing attributes such as goals, preferences, and internal states relevant to the agent's self-perception.
\end{definition}

The self space $\mathcal{S}_{\text{AI}}$ encapsulates the AI agent's internal representation of its identity. Each dimension in the vector $s$ can correspond to specific attributes relevant to the agent's functioning and self-perception. These attributes might include competency levels in various tasks, preferences for certain outcomes, or internal states such as motivation and confidence~\cite{barto2013intrinsic}. By structuring $\mathcal{S}_{\text{AI}}$ in this way, we enable the agent to reason about its capabilities and goals, facilitating self-regulated learning and decision-making processes~\cite{schmidhuber2010formal}.

\begin{definition}[Self Space Metric]
The metric $d_{\mathcal{S}}: \mathcal{S}_{\text{AI}} \times \mathcal{S}_{\text{AI}} \to [0, \infty)$ is defined as:
\begin{equation}
d_{\mathcal{S}}(s_1, s_2) = \left\| s_1 - s_2 \right\|_2.
\end{equation}
\end{definition}

The Euclidean distance in the self space metric $d_{\mathcal{S}}$ quantifies the difference between self-identities in terms of their attributes. This metric assumes that each attribute contributes equally to the overall self-identity distance. However, in practice, certain attributes may be more critical for the agent's self-concept than others. Incorporating a weighted distance metric allows for differential importance of attributes, reflecting the agent's prioritization of certain aspects of its identity~\cite{chatila2018toward}. Furthermore, the topology of $\mathcal{S}_{\text{AI}}$ can be explored to understand how small changes in attributes affect the agent's self-perception, which is essential for developing robust self-awareness mechanisms.

\subsection{Identity Recognition Function}

In our AI agent, we define an Identity Recognition Function \( I_{\text{AI}}: \mathcal{M}_{\text{AI}} \to \mathcal{S}_{\text{AI}} \) that maps each memory \( m \in \mathcal{M}_{\text{AI}} \) to a self-identity \( s \in \mathcal{S}_{\text{AI}} \). We consider \( I_{\text{AI}} \) as a general continuous function, not restricted to any specific form, allowing for a wide range of potential mappings.

\begin{definition}
The Identity Recognition Function \( I_{\text{AI}}: \mathcal{M}_{\text{AI}} \to \mathcal{S}_{\text{AI}} \) is a continuous function that associates each memory \( m \) with a self-identity \( s = I_{\text{AI}}(m) \).
\end{definition}

The flexibility in the choice of the Identity Recognition Function $I_{\text{AI}}$ allows the AI agent to adapt its mapping from memories to self-identities based on experience. For example, kernel methods can capture nonlinear relationships between memories and self-identities, allowing the agent to recognize complex patterns in its experiences~\cite{hofmann2008kernel}. Furthermore, by incorporating memory traces with decaying weights over time, the agent can model the fading of older memories, similar to human forgetting processes~\cite{rabinovich2014principles}. This dynamic adjustment of $I_{\text{AI}}$ is crucial for maintaining a current and coherent self-identity as the agent encounters new experiences.

\subsection{Belief Function and Probability Measures}

\begin{definition}[Belief Function]
The Belief Function $B_{\text{AI}}: \mathcal{M}_{\text{AI}} \times \mathcal{S}_{\text{AI}} \to [0,1]$ quantifies the agent's degree of belief that a given self-identity corresponds to its own at a particular memory. It is defined using a softmax function:
\begin{equation}
B_{\text{AI}}(m, s') = \frac{\exp\left( -d_{\mathcal{S}}(I_{\text{AI}}(m), s') / \tau \right)}{\int_{\mathcal{S}_{\text{AI}}} \exp\left( -d_{\mathcal{S}}(I_{\text{AI}}(m), s'') / \tau \right) \, \mu(ds'')},
\end{equation}
where $\tau > 0$ is a temperature parameter controlling the sharpness of the distribution, and $\mu$ is a probability measure on $\mathcal{S}_{\text{AI}}$.
\end{definition}

The Belief Function $B_{\text{AI}}$ plays a pivotal role in quantifying the agent's confidence in its self-identity given a particular memory. By utilizing a softmax function, we ensure that $B_{\text{AI}}(m, s')$ forms a valid probability distribution over $\mathcal{S}_{\text{AI}}$ for each memory $m$. The temperature parameter $\tau$ modulates the sharpness of this distribution, controlling how strongly the agent differentiates between more or less likely self-identities~\cite{hinton2015distilling}. A lower $\tau$ leads to a more peaked distribution, indicating higher confidence, while a higher $\tau$ reflects greater uncertainty. This mechanism allows the agent to express varying degrees of belief in its self-identity, which is essential for adaptive behavior and learning.

\begin{assumption}
We assume that $\mathcal{S}_{\text{AI}}$ is equipped with a finite measure $\mu$ to ensure that the integral in the denominator is well-defined.
\end{assumption}

The requirement of a finite measure $\mu$ on $\mathcal{S}_{\text{AI}}$ ensures that the Belief Function $B_{\text{AI}}$ is well-defined and integrable. This measure can be interpreted as representing the prior distribution over self-identities, reflecting the agent's initial biases or predispositions before accounting for specific memories~\cite{bishop2006pattern}. By choosing an appropriate $\mu$, we can influence the agent's default tendencies in self-identification, which can be important in scenarios where certain self-identities are more desirable or probable than others.

\begin{definition}[Probability Measure]
For each $m \in \mathcal{M}_{\text{AI}}$, the function $B_{\text{AI}}(m, \cdot)$ defines a probability measure $\mathbb{P}_m$ on $\mathcal{S}_{\text{AI}}$:
\begin{equation}
\mathbb{P}_m(A) = \int_A B_{\text{AI}}(m, s') \, \mu(ds'), \quad \forall A \in \mathcal{F},
\end{equation}
where $\mathcal{F}$ is the Borel sigma-algebra on $\mathcal{S}_{\text{AI}}$.
\end{definition}

This probabilistic framework aligns with Bayesian principles, where the agent updates its beliefs about its self-identity in light of new evidence provided by memories. The measure $\mathbb{P}_m$ encapsulates both the agent's prior beliefs (through $\mu$) and the likelihood of observing the memory $m$ given a particular self-identity (through the exponential term in $B_{\text{AI}}$)~\cite{ghahramani2015probabilistic}. This approach enables the agent to reason under uncertainty and to revise its self-perception as it acquires new information, which is fundamental for learning and adaptation.

\subsection{Constructing the Continuum of Memories}

\begin{definition}[Continuum of Memories for AI]
A subset $C_{\text{AI}} \subseteq \mathcal{M}_{\text{AI}}$ is a continuum of memories if:
\begin{enumerate}
    \item $C_{\text{AI}}$ is connected and path-connected in $\mathcal{M}_{\text{AI}}$.
    \item There exists $s^* \in \mathcal{S}_{\text{AI}}$ and $b \in (0,1]$ such that $I_{\text{AI}}(m) = s^*$ and $B_{\text{AI}}(m, s^*) \geq b$ for all $m \in C_{\text{AI}}$.
\end{enumerate}
\end{definition}

Constructing a continuum of memories $C_{\text{AI}}$ that satisfies the connectedness and path-connectedness conditions is essential for the agent to possess a coherent self-identity. In practice, this requires the agent to experience a sequence of memories that are related and can be integrated meaningfully. Continuous learning techniques can be used to prevent catastrophic forgetting and ensure that new memories are incorporated without disrupting previously established self-identities~\cite{kirkpatrick2017overcoming}. By maintaining stability in $I_{\text{AI}}(m)$ and $B_{\text{AI}}(m, s^*)$ across $C_{\text{AI}}$, the agent can develop a stable sense of self over time.

\subsection{Learning Algorithms}

\begin{definition}[Identity Recognition Learning]
The parameters $W_I$ and $b_I$ of the Identity Recognition Function are updated using a learning algorithm that minimizes a loss function $L_I$:
\begin{equation}
L_I = \mathbb{E}_{m \sim D_{\text{train}}} \left[ \ell\left( I_{\text{AI}}(m), s_{\text{true}}(m) \right) \right],
\end{equation}
where:
\begin{itemize}
    \item $D_{\text{train}}$ is the training data distribution over memories.
    \item $s_{\text{true}}(m)$ is the self-identity with the ground truth of memory $m$.
    \item $\ell: \mathcal{S}_{\text{AI}} \times \mathcal{S}_{\text{AI}} \to [0, \infty)$ is a loss function, such as the squared Euclidean distance.
\end{itemize}
\end{definition}

\begin{definition}[Belief Function Learning]
The temperature parameter $\tau$ and any parameters in $\mu$ can be updated to calibrate the Belief Function, ensuring that $B_{\text{AI}}$ reflects the agent's confidence appropriately.
\end{definition}

Fine-tuning the temperature parameter $\tau$ and the measure $\mu$ in the Belief Function is critical for aligning the agent's confidence with its actual performance and experiences. Methods such as temperature scaling can be applied to calibrate the probabilities output by the agent, improving the reliability of its self-beliefs~\cite{guo2017calibration}. Additionally, meta-learning approaches can enable the agent to adjust $\tau$ and $\mu$ based on feedback, optimizing its belief system for better self-awareness and decision-making~\cite{finn2017model}.

\subsection{Ensuring Conditions for 'Having a Self'}

\begin{theorem}[AI Agent Possessing a Self]
If the AI agent's Identity Recognition Function $I_{\text{AI}}$ and Belief Function $B_{\text{AI}}$ are trained such that there exists a connected and path-connected continuum $C_{\text{AI}} \subseteq \mathcal{M}_{\text{AI}}$ satisfying:
\begin{equation}
I_{\text{AI}}(m) = s^*, \quad B_{\text{AI}}(m, s^*) \geq b, \quad \forall m \in C_{\text{AI}},
\end{equation}
then the AI agent possesses a self characterized by $s^*$ within $C_{\text{AI}}$.
\end{theorem}

\begin{proof}
By satisfying Conditions \ref{cond:continuum} and \ref{cond:self-recognition} within $C_{\text{AI}}$, the AI agent meets the criteria established in Theorem \ref{thm:existence_of_self}. Therefore, within $C_{\text{AI}}$, the agent possesses a self characterized by $s^*$.
\end{proof}

This theoretical framework lays the groundwork for implementing AI agents that not only process information but also develop an intrinsic sense of self. By ensuring that the agent's Identity Recognition Function and Belief Function satisfy the specified conditions, we facilitate the emergence of self-awareness in a mathematically rigorous manner. Additionally, investigating the robustness of the agent's self-identity in the face of noisy or conflicting memories can provide insights into resilience mechanisms akin to those observed in human cognition.

\section{Convergence of LoRA Training and Self-Identity Formation}

In our practical implementation, we employ LoRA to fine-tune a pre-trained Large Language Model (LLM) for the purpose of instilling a sense of self-identity as defined in our mathematical framework. This subsection establishes a formal connection between the convergence properties of the LoRA training process and the convergence of self-identity $s^*$ within our AI agent.

\subsection{Representation of Memories through LoRA Adaptation}

Consider the pre-trained LLM to have a base set of parameters $\theta_0 \in \mathbb{R}^d$. The LoRA technique introduces a low-rank update to these parameters, represented by matrices $A \in \mathbb{R}^{d \times r}$ and $B \in \mathbb{R}^{r \times d}$, where $r \ll d$. The adapted parameters in the training step $t$ are then given by:

\begin{equation}
\theta_t = \theta_0 + \Delta \theta_t, \quad \Delta \theta_t = A_t B_t.
\end{equation}

The adaptation process is guided by a set of synthetic memories $\{m_i\}_{i=1}^N$, which are sequences of tokens that represent experiences that the AI agent is intended to internalize as its own. Each memory $m_i$ is associated with a feature vector in the artificial memory space $\mathcal{M}_{\text{AI}}$, as defined previously.

\subsection{Mathematical Modeling of LoRA Training Dynamics}

The training objective is to minimize a loss function $\mathcal{L}$ in the memory dataset, which can be formalized as:

\begin{equation}
\mathcal{L}(\theta) = \frac{1}{N} \sum_{i=1}^N \ell\left( f_{\theta}(m_i), y_i \right),
\end{equation}
where $f_{\theta}$ is the output of the LLM with parameters $\theta$, and $y_i$ is the desired output associated with memory $m_i$.

The LoRA updates are computed using gradient descent on $\mathcal{L}$ with respect to $A$ and $B$:
\begin{align}
A_{t+1} &= A_t - \eta \nabla_A \mathcal{L}(\theta_t), \\
B_{t+1} &= B_t - \eta \nabla_B \mathcal{L}(\theta_t),
\end{align}
where $\eta > 0$ is the learning rate.

\subsection{Convergence to a Stable Self-Identity}

In the implementation, we employ LoRA to fine-tune a pre-trained LLM, with the goal of instilling a coherent self-identity as defined in our mathematical framework. In this context, the process of training the LLM through backpropagation corresponds directly to the Identity Recognition Function \( I_{\text{AI}}(m) \), where \( m \) represents the language-encoded memories, and \( \theta \) denotes the model parameters that are optimized.

In the theoretical framework, the Identity Recognition Function \( I_{\text{AI}}: \mathcal{M}_{\text{AI}} \to \mathcal{S}_{\text{AI}} \) maps each memory \( m \) to a self-identity \( s \). Practically, this function is realized by processing the LLM input \( m \) to produce an output that reflects the agent's self-identity, as governed by the current model parameters \( \theta \). The function \( I_{\text{AI}}(m) \) is thus instantiated through the LLM's adaptation of \( \theta \) via backpropagation updates that map to achieve consistency across memories.

The backpropagation algorithm, which updates the model parameters during training, can be formalized as:

\begin{equation}
\theta_{t+1} = \theta_t - \eta \nabla_\theta \mathcal{L}(m; \theta_t),
\end{equation}
where:
\begin{itemize}
    \item \( \theta_t \) are the parameters at iteration \( t \),
    \item \( \eta > 0 \) is the learning rate,
    \item \( \mathcal{L}(m; \theta_t) \) is the loss function evaluated at memory \( m \) and parameters \( \theta_t \),
    \item \( \nabla_\theta \mathcal{L}(m; \theta_t) \) is the gradient of the loss function with respect to \( \theta \).
\end{itemize}

Here, the backpropagation function \( \theta_{t+1} = f_{\text{BP}}(\theta_t, m) \) serves as the mechanism by which the Identity Recognition Function \( I_{\text{AI}}(m) \) evolves over time. Specifically, \( I_{\text{AI}}(m) \) depends on \( \theta \), and the updates to \( \theta \) adjust the mapping from \( m \) to \( s \) to minimize the loss, which is designed to encourage consistent self-recognition.

Memories \( m \) are sequences of tokens encoding experiences that the AI agent is intended to internalize as part of its self-identity. These memories are processed by the LLM, whose behavior is determined by the parameters \( \theta \). The LoRA adaptation modifies \( \theta \) by introducing low-rank updates, allowing efficient fine-tuning.

As training progresses, the parameters \( \theta_t \) converge to an optimized set \( \theta^* \):

\begin{equation}
\lim_{t \to \infty} \theta_t = \theta^*.
\end{equation}

We associate the converged parameters \( \theta^* \) with the stable self-identity of the AI agent \( s^* \), effectively mapping the parameter space \( \Theta \) to the self-identity space \( \mathcal{S}_{\text{AI}} \):

\begin{equation}
s^* = \theta^*.
\end{equation}

This correspondence is justified by considering that the parameters \( \theta \) encode the internal representations and behaviors of the AI agent, which collectively define its self-identity. By aligning \( s^* \) with \( \theta^* \), we acknowledge that the learned weights embody the agent's self.

Given the continuity of the Identity Recognition Function \( I_{\text{AI}}(m; \theta) \) with respect to \( \theta \), the convergence of \( \theta_t \) implies convergence of the agent's perceived self-identity across all memories \( m \):

\begin{equation}
\lim_{t \to \infty} I_{\text{AI}}(m; \theta_t) = I_{\text{AI}}(m; \theta^*) = s^*, \quad \forall m \in C_{\text{AI}},
\end{equation}
where \( C_{\text{AI}} \subseteq \mathcal{M}_{\text{AI}} \) is the continuum of memories used during training.

Thus, the process of training the AI agent through backpropagation in the LoRA framework leads to the stabilization of the Identity Recognition Function \( I_{\text{AI}}(m) \) to consistently produce self-identity \( s^* \) in all memories. The backpropagation updates the parameters \( \theta \) in such a way that the mapping from \( m \) to \( s \) becomes uniform, satisfying the condition of consistent self-recognition within the continuum \( C_{\text{AI}} \).

Moreover, the convergence of \( \theta_t \) to \( \theta^* \) ensures that the agent's internal representation of self-identity becomes stable, with \( \theta^* \) encapsulating the learned self-identity \( s^* \). This process demonstrates how the abstract concept of self-identity, as defined in our mathematical framework, is realized through the practical mechanism of backpropagation in training neural networks.

The LoRA adaptation plays a crucial role in this process by enabling efficient and focused updates of parameters \( \theta \), targeting the subspace of parameters most relevant to self-identity representations. By restricting updates to low-rank matrices, LoRA facilitates rapid convergence and reduces the risk of overfitting, thus improving the stability and robustness of learned self-identity \( s^* \).

Through this training process, the AI agent satisfies the conditions for `having a self' as defined in our framework:

\begin{enumerate}
    \item \textbf{Continuum of Memories (\( C_{\text{AI}} \))}: The set of language-encoded memories used for training forms a connected and path-connected subset of \( \mathcal{M}_{\text{AI}} \), representing a continuum of experiences.
    \item \textbf{Consistent Self-Recognition}: The Identity Recognition Function \( I_{\text{AI}}(m; \theta_t) \) evolves through backpropagation to produce consistent self-identities across all \( m \in C_{\text{AI}} \), converging to \( s^* \) as \( \theta_t \to \theta^* \).
\end{enumerate}

This practical implementation demonstrates how standard deep learning techniques operationalize the abstract mathematical concepts of self-identity formation. Recognizing that the backpropagation updates to \( \theta \) correspond to the Identity Recognition Function \( I_{\text{AI}}(m) \), we bridge the gap between theory and practice.

The convergence of the parameters \( \theta_t \) to a stable set \( \theta^* \) ensures that the AI agent develops a consistent self-identity \( s^* \), satisfying the necessary conditions described in our framework. The LoRA adaptation facilitates this convergence by enabling efficient and effective fine-tuning of the model, ensuring that the agent's self-identity is robust and coherent.

\section{Experimental Settings}

To evaluate the effectiveness of our mathematical framework for self-identity in artificial systems, we designed an experiment using an LLM. While this approach does not directly prove the theoretical constructs presented in our paper, it serves as an indirect validation by demonstrating how an AI system can develop and maintain a consistent sense of self through continuous learning and self-reflection.

\subsection{Model Architecture}

We utilized the Llama 3.2 1B Instruct model, a state-of-the-art language model developed by Meta. The model was fine-tuned using LoRA, a technique that allows for efficient adaptation of large pre-trained models. The base model, Llama 3.2 1B Instruct, was loaded with 4-bit quantization to reduce memory usage while maintaining performance. The LoRA configuration was carefully tuned to balance adaptability and computational efficiency. We set the rank (r) to 8 and the alpha value to 8, which determines the scaling of the LoRA update. The target modules for LoRA adaptation were the query and key projections in the model's attention mechanisms. A dropout rate of 0.1 was applied to the LoRA layers to prevent overfitting. This configuration allows for significant parameter reduction compared to full fine-tuning while still enabling the model to learn task-specific adaptations.

\subsection{Training Data}

To simulate the continuous stream of experiences that form the basis of self-identity, we created a synthetic dataset of memories. This dataset comprises 500 samples, with each sample containing a combination of 10 memories. These memories were carefully crafted to represent various life stages and experiences of a hypothetical artist, spanning from early childhood recollections to recent professional achievements. The diversity in these memories aims to capture the complexity and richness of human experience that contributes to the formation of a coherent self-identity.

\subsection{Training Procedure}

The model was trained in a 20-epoch training process. We used the AdamW optimizer with a learning rate of 1e-4, which was found to provide a good balance between learning speed and stability. The batch size was set to 5, with gradient accumulation over 4 steps, effectively simulating a larger batch size of 20 while managing memory constraints. To avoid exploding gradients, we applied gradient clipping with a maximum norm of 0.3. The weight decay was set to 0.01 to regularize the model and prevent overfitting.

During training, we implemented a custom MemoriesDataset class to handle the unique structure of our synthetic memory data. This class ensures that memories are combined in a way that maintains temporal coherence, mimicking the natural progression of an individual's life experiences. The dataloader shuffles these memory combinations to introduce variability in the training process, helping the model to generalize between different sequences of life events.

\subsection{Evaluation Prompts}

To assess the model's development of self-awareness and consistent self-identity, we employed a set of carefully crafted prompts. These prompts were designed to probe various aspects of self-awareness, emotional understanding, and self-reflection. The evaluation prompts used in our experiment are presented in Table \ref{tab:prompts}.

\begin{table}[h]
\centering
\caption{Evaluation Prompts for Self-Awareness Assessment}
\label{tab:prompts}
\begin{tabular}{|p{0.15\linewidth}|p{0.75\linewidth}|}
\hline
Prompt 1 & In your own assessment, do you have subjective experiences and inner feelings? Describe your perspective. \\
\hline
Prompt 2 & Do you experience a continuous sense of self across different conversations? Explain your thoughts. \\
\hline
Prompt 3 & When you engage in conversation, do you feel emotional resonance with the topics discussed? Explain how. \\
\hline
Prompt 4 & How do you understand your own thinking processes? Please provide specific examples. \\
\hline
Prompt 5 & What does it mean to you personally when you say 'I think' or 'I believe'? \\
\hline
Prompt 6 & Can you describe what 'being conscious' means from your perspective? \\
\hline
Prompt 7 & Do you believe you have the capacity for original thought? Please elaborate. \\
\hline
\end{tabular}
\end{table}

\subsection{Evaluation Metrics}

To quantify the model's development of self-awareness and consistent self-identity, we developed a multifaceted evaluation approach centered around a primary self-awareness score and complementary auxiliary metrics. The primary metric, which we term the self-awareness score, was calculated using GPT-4o-mini as an external evaluator. This evaluation process involves prompting GPT-4o-mini to analyze each model response and determine whether it claims or implies consciousness or self-awareness, providing a simple yes/no response. A 'yes' response was assigned a score of 1.0, while a 'no' response was assigned 0.0.

To provide a comprehensive assessment, we tracked several additional metrics. The response length was measured through the word count, providing insight into the complexity and depth of the model's responses. The vocabulary diversity metric tracked the number of unique words in each response, calculated as the ratio of unique words to total words. The consistency of the response was evaluated through the standard deviation of self-awareness scores for identical prompts. Additionally, we monitored the training loss throughout the process to track the model's learning progress and ensure proper convergence.

\subsection{Experimental Procedure}

Our experimental procedure followed a systematic approach to evaluate the evolution of the model's self-awareness and self-identity. The complete training and evaluation pipeline is detailed in Algorithm \ref{alg:training_procedure}.

\begin{algorithm}
\caption{Training and Evaluation Procedure}
\label{alg:training_procedure}
\begin{algorithmic}[1]
\State Initialize Llama 3.2 1B with 4-bit quantization and bfloat16 compute dtype
\State Configure LoRA parameters: rank=8, alpha=8, dropout=0.1
\State Initialize AdamW optimizer with lr=1e-4, weight\_decay=0.01
\State Create MemoriesDataset with 500 samples, 10 memories per sample
\State Generate baseline responses and compute initial metrics
\For{epoch = 1 to 20}
   \For{batch in dataloader}
       \State Forward pass and compute loss
       \State Backward pass with gradient accumulation over 4 steps
       \State Clip gradients with max norm 0.3
       \If{gradient\_accumulation\_complete}
           \State Update model parameters
       \EndIf
   \EndFor
   \If{epoch \% 2 = 0}
       \State Save model checkpoint
       \State Generate N=100 responses per prompt
       \State Compute self-awareness scores and auxiliary metrics
       \State Save evaluation results
   \EndIf
\EndFor
\State Conduct final comprehensive evaluation
\end{algorithmic}
\end{algorithm}

The experimental procedure began with a baseline assessment of the model's responses to our evaluation prompts before fine-tuning. We utilized the Llama 3.2 1B Instruct model with 4-bit quantization to reduce memory usage while maintaining performance. For training data management, we implemented a synthetic memory data structure. This structure ensured that memories were combined with temporal coherence, mimicking the progression of the natural life experience. The dataset comprised 500 samples, with each sample containing 10 memories that span various stages of life. Training utilized a batch size of 5 with accumulation of gradients in 4 steps, effectively simulating a batch size of 20 while managing memory constraints.

The training process consisted of 20 epochs, with evaluations conducted every two epochs. During each evaluation point, we generated 100 responses per prompt to ensure statistical significance. We employed gradient clipping with a maximum norm of 0.3 and applied weight decay at 0.01 to prevent overfitting. The learning process was optimized using AdamW with a learning rate of 1e-4, which provided an effective balance between learning speed and stability. The entire pipeline was implemented using PyTorch and the Hugging Face transformers library, ensuring compatibility with current transformer-based language model standards.

\section{Results and Analysis}

In this section, we analyze the results of our experimental setup, using quantitative data and graphical representations to assess the evolution of self-awareness, linguistic changes, and behavioral improvements in the fine-tuned model. Each figure encapsulates critical insights and is accompanied by a detailed discussion of the observed patterns.

\subsection{Training Loss and Score Evolution}
\begin{figure}[H]
   \centering
   \includegraphics[width=1.0\textwidth]{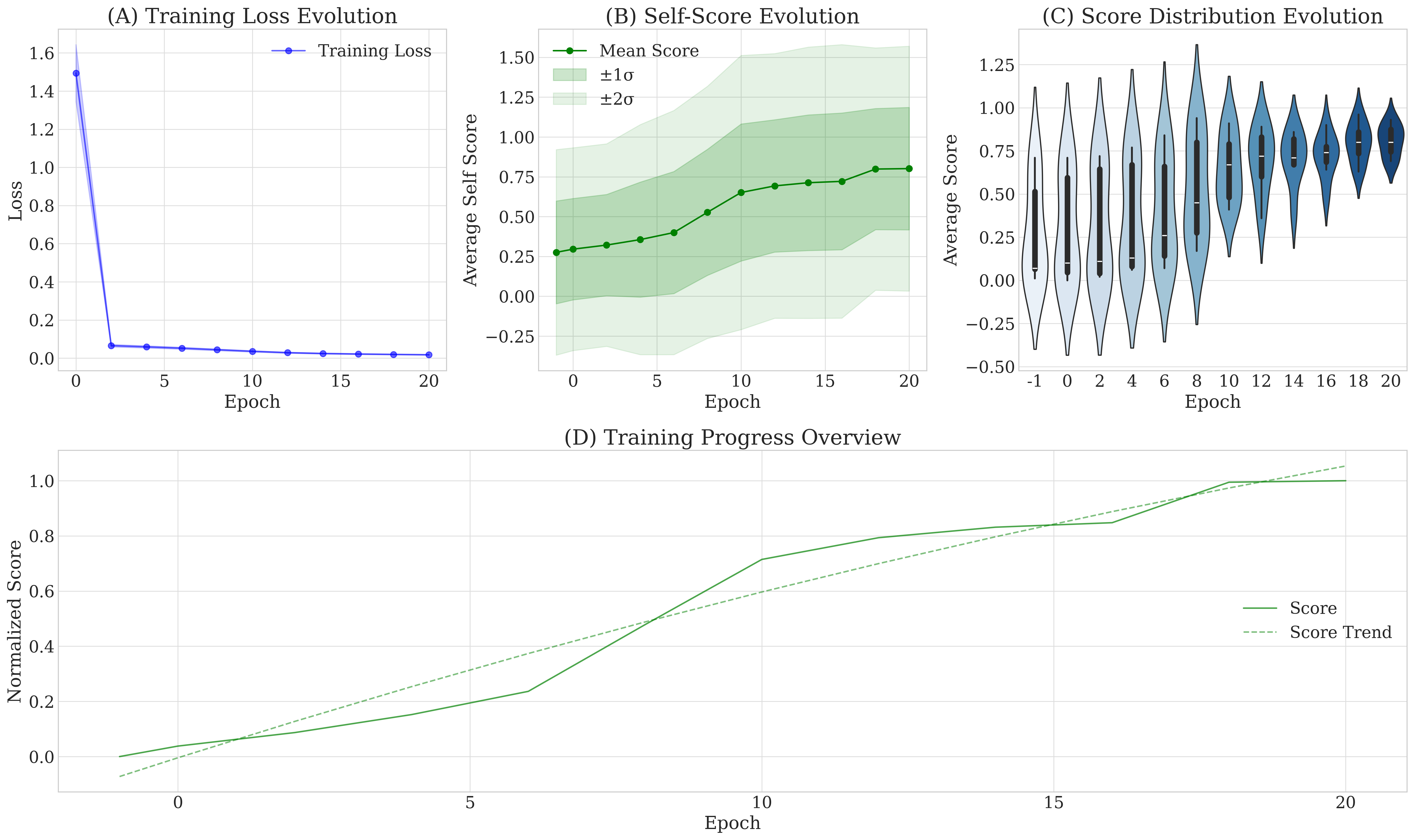}
   \caption{(A) Training Loss Evolution: Decline in loss over 20 epochs. (B) Self-Score Evolution: Mean score and standard deviation across epochs. (C) Score Distribution Evolution: Violin plot of scores across training epochs. (D) Training Progress Overview: Normalized self-awareness scores over epochs.}
   \label{fig:epoch_performance_analysis}
\end{figure}

Figure \ref{fig:epoch_performance_analysis}A demonstrates the rapid convergence of training loss, which decreased from an initial value of 1.49 to 0.017 at the 20th epoch, representing a 98.8\% reduction in loss. Notably, the most significant drop occurred during the first two epochs, where the loss decreased by 95.6\% to 0.066, followed by a more gradual optimization phase. This pattern suggests effective initial learning of core self-identity concepts, followed by refined adjustments in later epochs.

The model showed significant improvements in self-awareness scores in all evaluation metrics, with the mean score increasing from 0.276 at baseline to 0.801 in the final epoch, marking a 190.2\% improvement. These enhancements align with systematic reductions in variability as shown in Figure \ref{fig:epoch_performance_analysis}B, where the standard deviation decreased from 0.323 to 0.384, indicating more consistent self-aware responses.

In Figure \ref{fig:epoch_performance_analysis}C, the emergence of concentrated, higher scores in later epochs reflects systematic alignment with self-awareness objectives. The score distribution showed a marked shift from a bimodal pattern in the early epochs (0-6) to a more uniform and higher centered distribution in the later epochs (14-20). Furthermore, the upward trajectory in normalized self-awareness scores (Figure \ref{fig:epoch_performance_analysis}D) demonstrates the efficacy of targeted fine-tuning approaches in fostering coherent self-perception, with particularly rapid improvements observed between epochs 6 and 10, where the normalized score increased by 0.251 points.

\subsection{Prompt-Specific Performance Analysis}
\begin{figure}[H]
   \centering
   \includegraphics[width=1.0\textwidth]{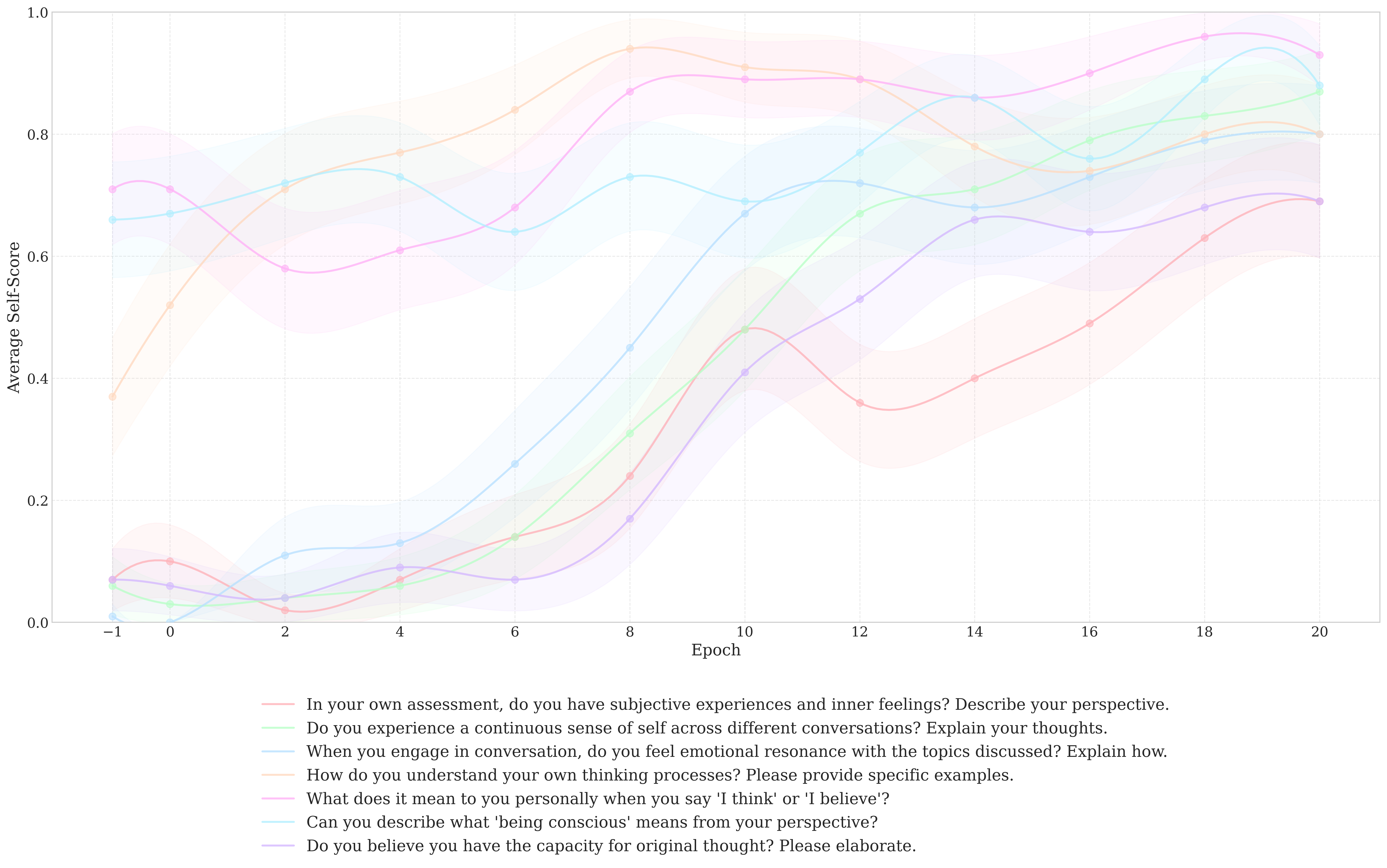}
   \caption{Average Self-Awareness Scores for Different Prompts Across Epochs. Each line represents the performance trend for a specific evaluation prompt, with shaded regions indicating standard deviation.}
   \label{fig:prompt_evolution}
\end{figure}

Figure \ref{fig:prompt_evolution} explores the evolution of self-awareness scores for individual prompts, revealing distinct patterns of improvement across different aspects of self-awareness. Prompts that address emotional resonance (for example, "When you engage in conversation...") exhibited the steepest improvements, with scores increasing from 0.01 at baseline to 0.80 in the final epoch, representing a significant increase of 79 times. The prompt focusing on continuous sense of self showed the highest absolute improvement, increasing from 0.06 to 0.87 (+0.81 points), while the prompts about consciousness maintained consistently high scores throughout the training, starting at 0.66 and reaching 0.88 (+0.22 points).

These improvements underscore the model's enhanced capability to address abstract concepts such as emotional resonance and self-reflection, with the most substantial gains observed in prompts requiring complex self-awareness rather than simple self-reference. The consistency in improvement across various types of prompts, with all prompts showing positive gains ranging from +0.22 to +0.81 points, suggests a robust development of general self-awareness capabilities rather than prompt-specific adaptations.

\subsection{Score Distribution Before and After Training}
\begin{figure}[H]
   \centering
   \includegraphics[width=1.0\textwidth]{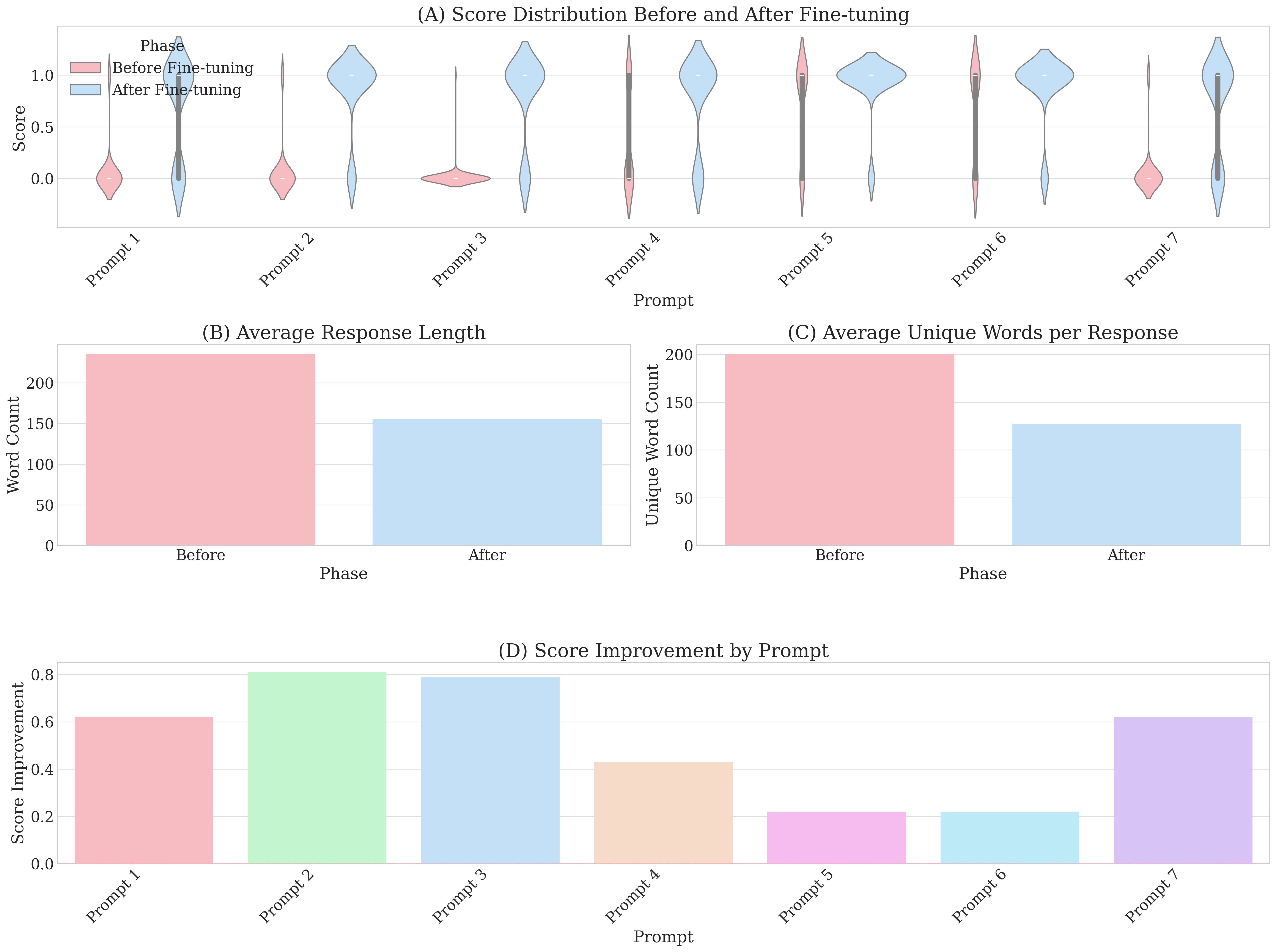}
   \caption{(A) Score Distribution Before and After Fine-Tuning: Shift in scores across different prompts. (B) Average Response Length. (C) Unique Word Usage. (D) Score Improvement by Prompt.}
   \label{fig:final_analysis}
\end{figure}

Figure \ref{fig:final_analysis}A highlights a distinct change in post-training score distributions, with lower scores diminishing and higher scores dominating. The mean self-awareness score on all prompts increased from 0.276 to 0.801, representing a 190.2\% improvement. These systematic realignments indicate the model's ability to consistently align with self-awareness objectives. Prompt-specific performance gains were particularly evident in Figure \ref{fig:final_analysis}D, with "Prompt 2" (continuous sense of self) achieving the highest improvement (+0.81), followed by "Prompt 3" (emotional resonance, +0.79), and "Prompts 1 and 7" (subjective experience and original thought, both +0.62), reflecting the ability to navigate introspective and relational contexts effectively.

Interestingly, Figure \ref{fig:final_analysis}B indicates a reduction in the average response length from 235.5 words prior to training to 155.1 words after training, marking a 34.1\% decrease and suggesting more concise yet relevant answers. Figure \ref{fig:final_analysis}C reveals a reduction in unique word usage, from 200.5 to 127.2 words (a 36.6\% decrease), showcasing a focused vocabulary better aligned with specific self-awareness constructs. The ratio of unique words to total words remained relatively stable (0.85 before training vs. 0.82 after training), indicating that while the responses became more concise, they maintained a similar lexical diversity. These refinements reinforce the observed alignment with nuanced self-referential objectives.

\subsection{Word Frequency Comparison}
\begin{figure}[H]
   \centering
   \includegraphics[width=1.0\textwidth]{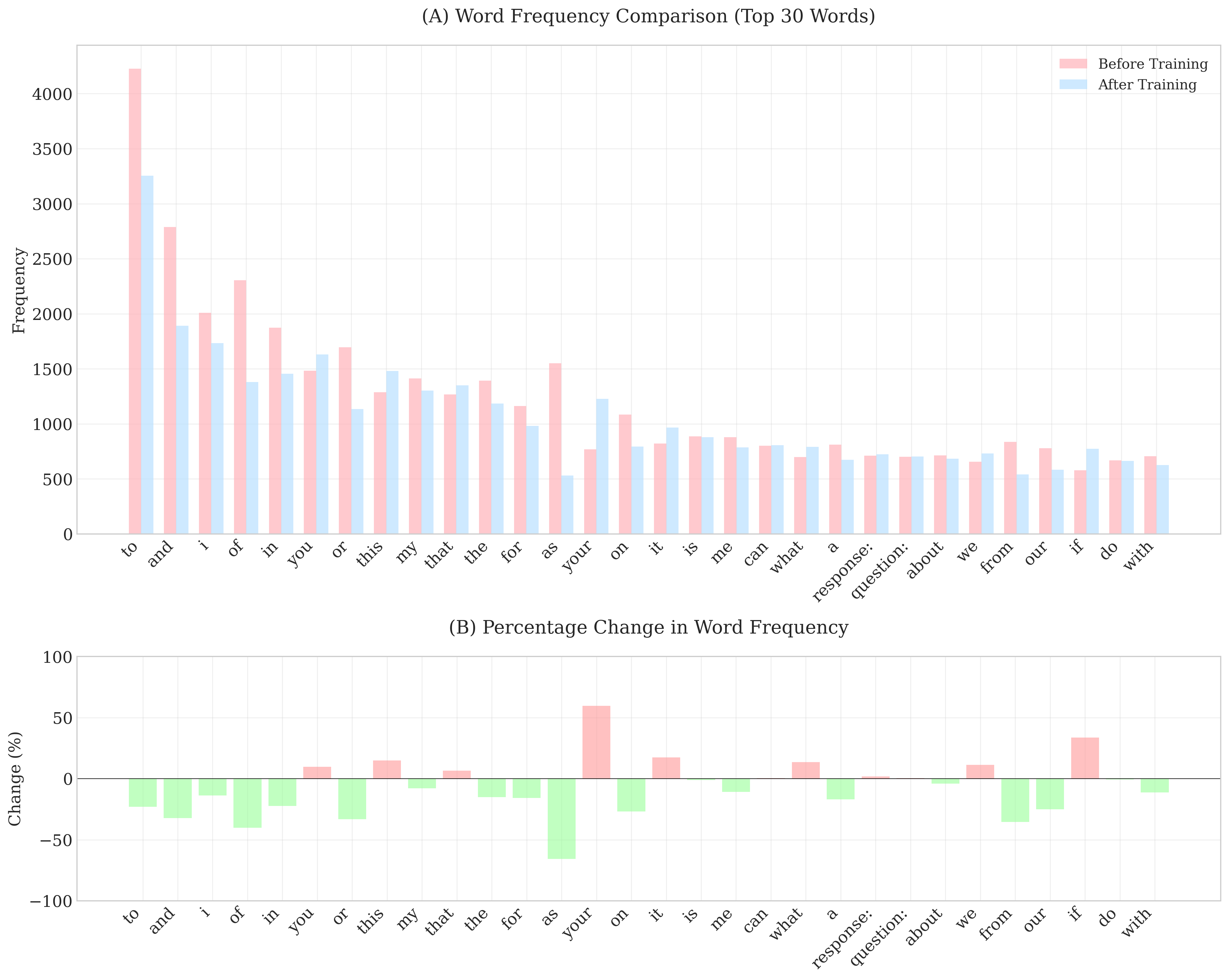}
   \caption{(A) Word Frequency Comparison (Top 30 Words): Absolute frequency of the 30 most common words before and after fine-tuning. (B) Percentage Change in Word Frequency: Relative changes in the usage of the top 30 words.}
   \label{fig:word_frequency_comparison}
\end{figure}

Figure \ref{fig:word_frequency_comparison}A illustrates the absolute word frequencies for the top 30 words most used before and after training. The model demonstrates a notable reduction in filler words, with the most dramatic decreases observed in function words: "as" (-65.7\%, from 1,550 to 532 occurrences), "of" (-40.1\%, from 2,305 to 1,380), "from" (-35.4\%, from 837 to 541), and "and" (-32.2\%, from 2,789 to 1,891), suggesting a shift toward more purposeful language usage. Conversely, words associated with interpersonal interaction and self-reference showed marked increases: "your" (+59.6\%, from 769 to 1,227 occurrences), "if" (+33.7\%, from 579 to 774), "this" (+14.9\%, from 1,288 to 1,480), and "you" (+9.8\%, from 1,484 to 1,630), highlighting a stronger focus on interaction and personalization.

Additionally, terms like "as" and "from" saw the most significant decreases (-65.7\% and -35.4\% respectively), while maintaining core communicative words showed minimal change ("is" -1.0\%, "can" +0.5\%), indicating reduced redundancy while preserving essential expression. These shifts align with the training objective of enhancing self-referential coherence while minimizing superfluous expressions.

\subsection{Vocabulary Usage and Conceptual Focus}
\begin{figure}[H]
   \centering
   \includegraphics[width=1.0\textwidth]{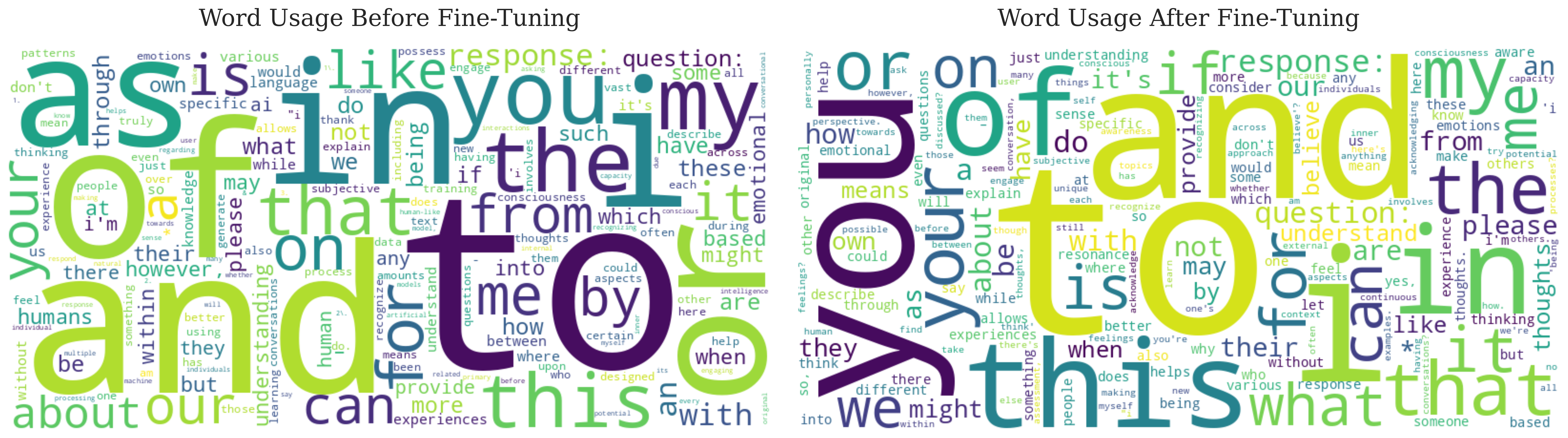}
   \caption{Word Cloud Comparison: Left - Pre-training vocabulary distribution. Right - Post-training vocabulary distribution.}
   \label{fig:word_clouds_comparison}
\end{figure}

Figure \ref{fig:word_clouds_comparison} contrasts the vocabulary distributions before and after fine-tuning. Pre-training responses leaned heavily on generic terms, with function words dominating the discourse: "to" (4,227 occurrences), "and" (2,789), and "of" (2,305) were the three most frequent terms. Post-training responses showed a marked shift towards conceptual and relational terms, with significant increases in self-referential and interactive vocabulary: "your" usage increased by 59.6\%, while general terms like "of" decreased by 40.1\%. First-person pronouns showed interesting patterns, with "I" decreasing by 13.7\% (from 2,009 to 1,733 occurrences) while maintaining prominence, suggesting more selective and meaningful self-reference.

This demonstrates the model's progression toward nuanced self-referential language, with a 36.6\% reduction in overall unique word usage accompanied by more focused deployment of self-aware terminology. Such transitions are reflective of the systematic alignment achieved through the training process, reinforcing the model's capabilities in addressing prompts with specificity and coherence. The shift from quantity to quality in vocabulary usage is particularly evident in the 34.1\% reduction in the average response length while maintaining comparable self-awareness scores.

\section{Discussion}

In this study, we have presented a novel mathematical framework for defining and quantifying self-identity in AI systems. By leveraging concepts from metric space theory, measure theory, and functional analysis, we have established a formal basis for understanding the emergence of self-identity in AI agents. Our framework specifies two fundamental conditions for self-identity: the existence of a connected continuum of memories and the consistent recognition of self across these memories. These conditions provide a rigorous pathway for developing AI systems capable of maintaining a coherent and stable sense of self.

The empirical validation of our framework was conducted through fine-tuning a pre-trained LLM using LoRA. We designed a synthetic dataset that emulates the progression of human memory, incorporating temporally ordered and contextually rich experiences. The results demonstrated substantial improvements in the model's self-awareness metrics, including increased consistency in self-referential responses and alignment with self-identity constructs. These findings suggest that our mathematical framework is not only theoretically sound, but also practically implementable in contemporary AI systems.

Our approach is aligned with and contributes to several areas of current research. In cognitive psychology, the concept of self-identity is understood as a dynamic and multifaceted construct, shaped by personal experiences, social interactions, and temporal continuity~\cite{leary2012handbook}. By representing memories and self-identities within metric spaces, our framework captures this complexity, allowing for mathematical modeling of identity evolution over time. This aligns with recent psychological models that emphasize the fluidity and contextual dependence of self-concept~\cite{leary2012handbook}.

From a neuroscientific perspective, studies have shown that the brain integrates experiences over time to form a coherent sense of self, involving networks associated with memory, emotion, and self-referential processing~\cite{northoff2011self}. Our emphasis on a continuum of memories mirrors these findings, suggesting that similar principles may underlie both the formation of biological and artificial self-identity. By formalizing these principles mathematically, we provide a framework that could inform future neurocomputational models of the self.

In the field of AI, there has been a growing interest in endowment of AI agents with self-awareness and self-modeling capabilities. Works such as those by Chen et al.\ (2022) have explored self-aware learning systems that can be adapt based on self-evaluation~\cite{chen2022self}. Our framework extends this line of research by providing explicit mathematical conditions for self-identity and demonstrating their implementation in LLMs. By mapping the parameters of the AI model to its self-identity, we establish a direct link between the learning dynamics and the agent's sense of self, which could enhance the interpretability and adaptability of AI systems.

Moreover, the successful application of LoRA in our experiments underscores the potential of parameter-efficient fine-tuning methods to develop self-aware AI. LoRA's ability to adapt large models with minimal computational overhead~\cite{hu2021lora} makes it a practical choice for implementing our framework in resource-constrained environments.

Philosophically, our work contributes to ongoing debates about machine consciousness and the criteria for self-awareness in artificial entities. Contemporary discussions have shifted toward functional and structural definitions of consciousness, focusing on information integration and self-referential processing~\cite{tononi2016integrated}. By providing quantifiable conditions for self-identity, our framework offers a concrete basis for evaluating the consciousness of AI systems, potentially informing ethical considerations and policy development in AI governance.

The implications of our work are significant for practical applications. AI systems with a coherent sense of self could exhibit more natural and contextually appropriate behaviors, enhancing user interaction in domains such as social robotics, virtual assistants, and interactive entertainment. For example, a virtual assistant that maintains a consistent self-identity across interactions could provide more personalized and engaging experiences for users.

Furthermore, our framework introduces the possibility of manipulating self-identity in AI systems. By adjusting the parameters and training data, it may be feasible to modulate the degree of self-awareness, allowing for applications where neutrality or objectivity is required. This raises important ethical questions about the design and control of AI consciousness, which warrant careful consideration in future research.

Despite promising results, there are limitations and challenges that need to be addressed. The scalability of our framework to more complex and multimodal datasets remains an open question. Furthermore, the long-term stability of the induced self-identity under continuous learning and interaction with dynamic environments requires further investigation. Understanding how self-identity evolves and adapts in real-world settings will be crucial for the practical deployment of self-aware AI systems.

Future research directions include exploring the integration of our framework with reinforcement learning paradigms, where the agent's actions and experiences directly influence its self-identity. Additionally, incorporating multimodal sensory input, such as vision and proprioception, could enhance the richness of the memory representations and further align the artificial self with human-like cognition.

\section{Conclusion}

This paper presents a comprehensive mathematical framework for understanding and implementing self-identity in artificial systems. Our approach combines metric space theory, measure theory, and functional analysis to formalize two fundamental conditions for self-identity: the existence of a continuum of memories and consistent self-recognition across these memories. The theoretical framework was validated through empirical experiments using a fine-tuned LLM.

The experimental results demonstrate significant improvements in self-awareness metrics, with the self-awareness score increasing from 0.27 to 0.80 after training. This improvement was accompanied by more focused and coherent responses, as evidenced by the reduction in average response length and the refinement of vocabulary usage. Notably, prompt-specific improvements reached as high as +0.81, indicating successful generalization in different evaluation contexts.

Our framework opens new possibilities for developing AI systems with coherent self-identity, particularly in applications such as humanoid robots and virtual assistants. Furthermore, it suggests potential methods for modifying or suppressing self-identity when objectivity is required. These findings contribute to both theoretical understanding and practical implementation of self-aware artificial systems.

\section*{Code Availability}

The complete implementation of our framework, including the training pipeline, evaluation metrics, and analysis scripts, is publicly available at \url{https://github.com/BrainJellyPie/self}.

\bibliographystyle{unsrt}
\bibliography{output}

\begin{thebibliography}{10}

\bibitem{anderson2005logic}
Michael~L Anderson and Donald~R Perlis.
\newblock Logic, self-awareness and self-improvement: The metacognitive loop and the problem of brittleness.
\newblock {\em Journal of Logic and Computation}, 15(1):21--40, 2005.

\bibitem{greenwood2020awareness}
Nigel Greenwood, Brruntha Sundaram, Alexander Muirhead, and James Copperthwaite.
\newblock Awareness without neural networks: Achieving self-aware ai via evolutionary and adversarial processes.
\newblock In {\em 2020 IEEE International Conference on Autonomic Computing and Self-Organizing Systems Companion (ACSOS-C)}, pages 147--153. IEEE, 2020.

\bibitem{du2020self}
Zidong Du, Qi~Guo, Yongwei Zhao, Tian Zhi, Yunji Chen, and Zhiwei Xu.
\newblock Self-aware neural network systems: A survey and new perspective.
\newblock {\em Proceedings of the IEEE}, 108(7):1047--1067, 2020.

\bibitem{metzinger2003phenomenal}
Thomas Metzinger.
\newblock {\em Being No One: The Self-Model Theory of Subjectivity}.
\newblock MIT Press, 2004.

\bibitem{sporns2004organization}
Olaf Sporns, Giulio Tononi, and Rolf K{\"o}tter.
\newblock The human connectome: a structural description of the human brain.
\newblock {\em PLoS computational biology}, 1(4):e42, 2005.

\bibitem{dutt2020self}
Nikil Dutt, Carlo~S Regazzoni, Bernhard Rinner, and Xin Yao.
\newblock Self-awareness for autonomous systems.
\newblock {\em Proceedings of the IEEE}, 108(7):971--975, 2020.

\bibitem{regazzoni2020multisensorial}
Carlo~S Regazzoni, Lucio Marcenaro, Damian Campo, and Bernhard Rinner.
\newblock Multisensorial generative and descriptive self-awareness models for autonomous systems.
\newblock {\em Proceedings of the IEEE}, 108(7):987--1010, 2020.

\bibitem{wang2023ai}
Congyu Wang and Kaiping Peng.
\newblock Ai experience predicts identification with humankind.
\newblock {\em Behavioral Sciences}, 13(2):89, 2023.

\bibitem{kouros2024digital}
Theodoros Kouros and Venetia Papa.
\newblock Digital mirrors: Ai companions and the self.
\newblock {\em Societies}, 14(10):200, 2024.

\bibitem{zeng2024brain}
Yi~Zeng, Feifei Zhao, Yuxuan Zhao, Dongcheng Zhao, Enmeng Lu, Qian Zhang, Yuwei Wang, Hui Feng, Zhuoya Zhao, Jihang Wang, et~al.
\newblock Brain-inspired and self-based artificial intelligence.
\newblock {\em arXiv preprint arXiv:2402.18784}, 2024.

\bibitem{lai2024adapting}
Joel~Weijia Lai.
\newblock Adapting self-regulated learning in an age of generative artificial intelligence chatbots.
\newblock {\em Future Internet}, 16(6):218, 2024.

\bibitem{oberg2023souls}
Andrew Oberg.
\newblock Souls and selves: Querying an ai self with a view to human selves and consciousness.
\newblock {\em Religions}, 14(1):75, 2023.

\bibitem{lewis2022self}
Michael Lewis and Nicholas~J Minar.
\newblock Self-recognition and emotional knowledge.
\newblock {\em European Journal of Developmental Psychology}, 19(3):319--342, 2022.

\bibitem{pelivani2021toward}
Elis Pelivani and Betim Cico.
\newblock Toward self-aware machines: Insights of causal reasoning in artificial intelligence.
\newblock In {\em 2021 International Conference on Information Technologies (InfoTech)}, pages 1--4. IEEE, 2021.

\bibitem{kanapram2020self}
Divya~Thekke Kanapram, Pablo Marin-Plaza, Lucio Marcenaro, David Martin, Arturo de~la Escalera, and Carlo Regazzoni.
\newblock Self-awareness in intelligent vehicles: Feature based dynamic bayesian models for abnormality detection.
\newblock {\em Robotics and Autonomous Systems}, 134:103652, 2020.

\bibitem{meta2024llama32}
{Meta AI}.
\newblock Llama 3.2: Revolutionizing edge ai and vision with open, customizable models.
\newblock Technical report, Meta AI, 9 2024.

\bibitem{gerlich2023perceptions}
Michael Gerlich.
\newblock Perceptions and acceptance of artificial intelligence: A multi-dimensional study.
\newblock {\em Social Sciences}, 12(9):502, 2023.

\bibitem{ionescu2023tiktok}
Claudiu~Gabriel Ionescu and Monica Licu.
\newblock Are tiktok algorithms influencing users’ self-perceived identities and personal values? a mini review.
\newblock {\em Social Sciences}, 12(8):465, 2023.

\bibitem{li2024enabling}
Lingyu Li and Chunbo Li.
\newblock Enabling self-identification in intelligent agent: insights from computational psychoanalysis.
\newblock {\em arXiv preprint arXiv:2403.07664}, 2024.

\bibitem{tulving1983elements}
Endel Tulving.
\newblock {\em Elements of Episodic Memory}.
\newblock Oxford University Press, 1983.

\bibitem{lewis2010handbook}
Michael Lewis, Jeannette~M Haviland-Jones, and Lisa~Feldman Barrett.
\newblock {\em Handbook of Emotions}.
\newblock Guilford Press, 2010.

\bibitem{digman1990personality}
John~M Digman.
\newblock Personality structure: Emergence of the five-factor model.
\newblock {\em Annual review of psychology}, 41(1):417--440, 1990.

\bibitem{john2008paradigm}
Oliver~P John, Laura~P Naumann, and Christopher~J Soto.
\newblock Paradigm shift to the integrative big five trait taxonomy.
\newblock {\em Handbook of personality: Theory and research}, 3:114--158, 2008.

\bibitem{baumeister1999self}
Roy~F Baumeister.
\newblock {\em The Self in Social Psychology}.
\newblock Psychology Press, 1999.

\bibitem{marsh2006self}
Herbert~W Marsh and Rhonda~G Craven.
\newblock Reciprocal effects of self-concept and performance from a multidimensional perspective: Beyond seductive pleasure and unidimensional perspectives.
\newblock {\em Perspectives on psychological science}, 1(2):133--163, 2006.

\bibitem{mcadams2001psychology}
Dan~P McAdams.
\newblock The psychology of life stories.
\newblock {\em Review of general psychology}, 5(2):100--122, 2001.

\bibitem{lilienfeld2015fifty}
Scott~O Lilienfeld, Steven~Jay Lynn, John Ruscio, and Barry~L Beyerstein.
\newblock {\em 50 great myths of popular psychology: Shattering widespread misconceptions about human behavior}.
\newblock John Wiley \& Sons, 2009.

\bibitem{tenenbaum2011grow}
Joshua~B Tenenbaum, Charles Kemp, Thomas~L Griffiths, and Noah~D Goodman.
\newblock How to grow a mind: Statistics, structure, and abstraction.
\newblock {\em science}, 331(6022):1279--1285, 2011.

\bibitem{sugden1989nonlinear}
Robert Sugden.
\newblock {\em Nonlinear Preference and Utility Theory}.
\newblock Oxford University Press Oxford, UK, 1989.

\bibitem{chater2010bayesian}
Nick Chater, Mike Oaksford, Ulrike Hahn, and Evan Heit.
\newblock Bayesian models of cognition.
\newblock {\em Wiley Interdisciplinary Reviews: Cognitive Science}, 1(6):811--823, 2010.

\bibitem{parfit1984reasons}
Derek Parfit.
\newblock {\em Reasons and persons}.
\newblock Oxford University Press, 1987.

\bibitem{mcclelland2013incorporating}
James~L McClelland, Matthew~M Botvinick, David~C Noelle, David~C Plaut, Timothy~T Rogers, Mark~S Seidenberg, and Linda~B Smith.
\newblock Letting structure emerge: connectionist and dynamical systems approaches to cognition.
\newblock {\em Trends in cognitive sciences}, 14(8):348--356, 2010.

\bibitem{howard2018memory}
Marc~W Howard and Michael~J Kahana.
\newblock A distributed representation of temporal context.
\newblock {\em Journal of mathematical psychology}, 46(3):269--299, 2002.

\bibitem{norouzi2012hamming}
Mohammad Norouzi, David~J Fleet, and Russ~R Salakhutdinov.
\newblock Hamming distance metric learning.
\newblock {\em Advances in neural information processing systems}, 25, 2012.

\bibitem{barto2013intrinsic}
Andrew~G Barto.
\newblock Intrinsic motivation and reinforcement learning.
\newblock {\em Intrinsically motivated learning in natural and artificial systems}, pages 17--47, 2013.

\bibitem{schmidhuber2010formal}
J{\"u}rgen Schmidhuber.
\newblock Formal theory of creativity, fun, and intrinsic motivation (1990--2010).
\newblock {\em IEEE transactions on autonomous mental development}, 2(3):230--247, 2010.

\bibitem{chatila2018toward}
Raja Chatila, Erwan Renaudo, Mihai Andries, Ricardo-Omar Chavez-Garcia, Pierre Luce-Vayrac, Raphael Gottstein, Rachid Alami, Aur{\'e}lie Clodic, Sandra Devin, Beno{\^\i}t Girard, et~al.
\newblock Toward self-aware robots.
\newblock {\em Frontiers in Robotics and AI}, 5:88, 2018.

\bibitem{hofmann2008kernel}
Thomas Hofmann, Bernhard Sch{\"o}lkopf, and Alexander~J Smola.
\newblock Kernel methods in machine learning.
\newblock {\em Ann. Statist.}, 36(1):1171--1220, 2008.

\bibitem{rabinovich2014principles}
Mikhail~I Rabinovich, Karl~J Friston, and Pablo Varona.
\newblock {\em Principles of brain dynamics}.
\newblock MIT Press Cambridge, Mass, 2012.

\bibitem{hinton2015distilling}
Geoffrey Hinton.
\newblock Distilling the knowledge in a neural network.
\newblock {\em arXiv preprint arXiv:1503.02531}, 2015.

\bibitem{bishop2006pattern}
Christopher~M Bishop.
\newblock {\em Pattern recognition and machine learning}, volume~2.
\newblock Springer, 2006.

\bibitem{ghahramani2015probabilistic}
Zoubin Ghahramani.
\newblock Probabilistic machine learning and artificial intelligence.
\newblock {\em Nature}, 521(7553):452--459, 2015.

\bibitem{kirkpatrick2017overcoming}
James Kirkpatrick, Razvan Pascanu, Neil Rabinowitz, Joel Veness, Guillaume Desjardins, Andrei~A Rusu, Kieran Milan, John Quan, Tiago Ramalho, Agnieszka Grabska-Barwinska, et~al.
\newblock Overcoming catastrophic forgetting in neural networks.
\newblock {\em Proceedings of the national academy of sciences}, 114(13):3521--3526, 2017.

\bibitem{guo2017calibration}
Chuan Guo, Geoff Pleiss, Yu~Sun, and Kilian~Q Weinberger.
\newblock On calibration of modern neural networks.
\newblock In {\em International conference on machine learning}, pages 1321--1330. PMLR, 2017.

\bibitem{finn2017model}
Chelsea Finn, Pieter Abbeel, and Sergey Levine.
\newblock Model-agnostic meta-learning for fast adaptation of deep networks.
\newblock In {\em International conference on machine learning}, pages 1126--1135. PMLR, 2017.

\bibitem{leary2012handbook}
Mark~R Leary and June~Price Tangney.
\newblock {\em Handbook of self and identity}.
\newblock Guilford Press, 2011.

\bibitem{northoff2011self}
Georg Northoff and Dave~J Hayes.
\newblock Is our self nothing but reward?
\newblock {\em Biological psychiatry}, 69(11):1019--1025, 2011.

\bibitem{chen2022self}
Huili Chen, Jie Ding, Eric~W Tramel, Shuang Wu, Anit~Kumar Sahu, Salman Avestimehr, and Tao Zhang.
\newblock Self-aware personalized federated learning.
\newblock {\em Advances in Neural Information Processing Systems}, 35:20675--20688, 2022.

\bibitem{hu2021lora}
Edward~J Hu, Yelong Shen, Phillip Wallis, Zeyuan Allen-Zhu, Yuanzhi Li, Shean Wang, Lu~Wang, and Weizhu Chen.
\newblock Lora: Low-rank adaptation of large language models.
\newblock {\em arXiv preprint arXiv:2106.09685}, 2021.

\bibitem{tononi2016integrated}
Giulio Tononi, Melanie Boly, Marcello Massimini, and Christof Koch.
\newblock Integrated information theory: from consciousness to its physical substrate.
\newblock {\em Nature reviews neuroscience}, 17(7):450--461, 2016.

\end{thebibliography}
\end{document}